\documentclass[]{beingbeyond}
\usepackage{enumitem}
\usepackage[toc,page,header]{appendix}

\usepackage[utf8]{inputenc} 
\usepackage[T1]{fontenc}    
\usepackage{hyperref}       
\usepackage{url}            
\usepackage{array}          
\usepackage{booktabs}       
\usepackage{amsfonts}       
\usepackage{nicefrac}       
\usepackage{microtype}      
\usepackage{xcolor}         
\usepackage{xspace}
\usepackage{bm}
\usepackage{bbm}
\usepackage{tabularx}
\usepackage{amssymb}
\usepackage{enumitem}
\usepackage{amsmath}
\usepackage{mathtools}
\usepackage{amsthm}
\usepackage{multirow}
\usepackage{makecell}
\usepackage{color}
\usepackage{colortbl}
\usepackage{adjustbox}
\usepackage{caption}
\usepackage{graphicx}
\usepackage{algorithm}
\usepackage{algpseudocode}
\usepackage{xspace}
\makeatletter
\DeclareRobustCommand\onedot{\futurelet\@let@token\@onedot}
\def\@onedot{\ifx\@let@token.\else.\null\fi\xspace}

\def\eg{\emph{e.g}\onedot}

\def\ie{\emph{i.e}\onedot}

\def\vs{\emph{vs}\onedot}

\newcommand{\ours}{Being-VL-0.5\xspace}

\title{Unified Multimodal Understanding via Byte-Pair Visual Encoding}

\author{{\bfseries Wanpeng Zhang$^{1,4}$ \ \ Yicheng Feng$^{1,4}$ \ \ Hao Luo$^{1,4}$ \ \ Yijiang Li$^{2}$\ \ Zihao Yue$^{3}$\\ Sipeng Zheng$^{4}$\ \ Zongqing Lu$^{1,4\dagger}$}}

\affiliation{{$^{1}$Peking University \quad $^{2}$UC San Diego \quad $^{3}$Renmin University of China \quad $^{4}$BeingBeyond}}

\webpage{\url{https://beingbeyond.github.io/Being-VL-0.5}}

\abstract{
Multimodal large language models (MLLMs) have made significant progress in vision-language understanding, yet effectively aligning different modalities remains a fundamental challenge. We present a framework that unifies multimodal understanding by applying byte-pair encoding to visual tokens. Unlike conventional approaches that rely on modality-specific encoders, our method directly incorporates structural information into visual tokens, mirroring successful tokenization strategies in text-only language models. We introduce a priority-guided encoding scheme that considers both frequency and spatial consistency, coupled with a multi-stage training procedure based on curriculum-driven data composition. These enhancements enable the transformer model to better capture cross-modal relationships and reason with visual information. Comprehensive experiments demonstrate improved performance across diverse vision-language tasks. By bridging the gap between visual and textual representations, our approach contributes to the advancement of more capable and efficient multimodal foundation models.
}

\begin{document}

\maketitle

\begingroup
\renewcommand\thefootnote{\fnsymbol{footnote}} 
\setcounter{footnote}{0}
\footnotetext[1]{Accepted by ICCV 2025.}
\footnotetext[2]{Correspondence to Zongqing Lu $<$lu@beingbeyond.com$>$.}
\endgroup

\section{Introduction}

Multimodal Large Language Models (MLLMs) have made significant progress in integrating information from various modalities\citep{yin2023survey, liu2024visual, yang2025embodiedbench}. Nevertheless, effectively representing visual information and aligning it seamlessly with texts still remains a core challenge \citep{baltruvsaitis2018multimodal, song2023bridge, zeng2024can}. Conventional approaches primarily generate visual embeddings using modality-specific encoders \citep{zhang2024vision, jin2023unified, feng2024videoorion} (such as CLIP \citep{radford2021learning}), yet recent research suggests that directly discretizing visual information can facilitate a more unified token representation \citep{lu2022unified, team2024chameleon, lu2024unified, zheng2024unicode}. In a recent study, researchers introduced a visual encoding scheme to vision modality~\citep{zhang2025bpe}, analogous to the byte-pair encoding (BPE) tokenization~\citep{sennrich2015neural, radford2019language, kozma2024theoretical} in LLMs. Previous work~\citep{zhang2025bpe} theoretically demonstrated that structurally integrating discrete tokens enhances information preservation, thereby enabling the Transformer models \citep{vaswani2017attention} to more effectively learn from two-dimensional visual data.  Although the preliminary results are promising, translating this theoretical framework into practical multimodal models poses significant challenges: (1) devising sophisticated token encoding strategies beyond simple frequency counting; (2) identifying data composition strategies that maximize visual-language alignment; and (3) designing effective training procedures to fully leverage these BPE-encoded visual tokens.  In this paper, we address these challenges by exploring practical implementations and optimization strategies for BPE-based visual tokenization. Our comprehensive analysis highlights key factors that significantly influence performance across modalities and proposes corresponding optimization.

Specifically, we introduce a priority-guided encoding scheme that accounts for both co-occurrence frequency and spatial consistency of visual patterns.  This novel approach generates semantically enriched visual tokens, thereby preserving the inherent structural information in images more effectively. Moreover, we propose data composition strategies that leverage a curriculum-based approach tailored for optimizing multimodal learning, explicitly designed to complement the visual BPE tokenization process. Unlike conventional visual encoders \citep{radford2021learning, zhai2023sigmoid}, which extract visual features through a single forward pass \citep{xu2021vitae, zhang2023vitaev2}, visual BPE tokenization generates hierarchical representations where encoded tokens progressively capture increasingly intricate visual patterns. Given this hierarchical property, an aligned learning curriculum is necessary; tokens must first acquire foundational semantic meanings through elementary visual-text associations before advancing toward more sophisticated visual reasoning tasks. Consequently, a multi-stage training procedure is further derived with stage-specific parameter freezing strategies: (1) embedding alignment with frozen language model parameters, (2)  selective fine-tuning of early transformer layers, and (3) full model fine-tuning. This incremental training approach facilitates more effective cross-modal knowledge transfer while maintaining the linguistic capabilities inherent to the foundation model.

We conduct extensive experiments across various benchmarks, consistently demonstrating that our proposed optimization strategies achieve substantial performance improvements compared to baseline approaches and prior studies. In addition to quantitative improvements, we provide detailed analyses of token patterns, computational efficiency, and qualitative improvements in visual comprehension capabilities.

Our contributions are summarized as follows: 
\begin{itemize}
    \item A discrete tokenization approach that preserves structural information in images through priority-guided encoding, providing unified visual representation for MLLMs.
    \item A progressive training strategy that combines stage-specific parameter freezing, allowing models to build better visual understanding.
    \item Empirical evidence across multiple benchmarks and extensive analysis that validates the effectiveness of our unified encoding methods and elucidates the reason why it advances visual-language understanding.
\end{itemize}
\section{Related Work}

\subsection{Representation Methods in MLLMs}

The core challenge of Multimodal Large Language Models (MLLMs) lies in effectively representing and integrating information from different modalities. The research community has primarily developed two representative approaches: methods based on continuous representations and methods based on discrete tokens.

Continuous token representation methods have emerged as the mainstream paradigm, with representative works such as LLaVA \citep{liu2024visual, liu2024improved}, Emu \citep{sun2023emu, wang2024emu3}, DeepSeek-VL \citep{lu2024deepseekvl,wu2024deepseek}, and Qwen-VL \citep{Qwen-VL, Qwen2-VL, Qwen2.5-VL}. These models use pre-trained visual encoders (\eg, CLIP \citep{radford2021learning}) to map images into continuous vector representations, which are then aligned with language models through projection layers \citep{zhang2023llama, gao2023llama}. Although this approach has achieved significant results across various downstream tasks, it faces two fundamental challenges: first, the modality gap between encoders and language models \citep{jiang2024specific,barua2023systematic}, wherein the high-dimensional continuous features provided by visual encoders differ from the discrete token representations expected by language models; second, the information bottleneck problem, wherein images compressed through visual encoders may lose substantial detailed information, particularly affecting low-frequency visual patterns \citep{rahmanzadehgervi2024vision}. Recent analysis \citep{bai2024hallucination,huang2024visual,ye2023cognitive} indicates that this architectural design makes models prone to visual hallucinations, generating descriptions even when relevant visual information is absent.

Discrete token representation methods have gained increasing attention in recent years, with representative works including  Unicode \citep{zheng2024unicode}, Unified-IO \citep{lu2022unified,lu2024unified}, and Chameleon \citep{team2024chameleon}. These methods typically use vector quantization models (such as VQ-GAN \citep{esser2021taming} or VQVAE \citep{van2017neural,razavi2019generating}) to discretize images into a series of tokens, thereby enabling visual information to be processed in a form similar to text. The advantage of this approach lies in its ability to process multimodal inputs in a unified manner, thus reducing inter-modal differences. However, existing discrete token methods also have significant limitations: most works simply quantize images into fixed-size tokens without considering the semantic structure of visual content; information loss during quantization may lead to the loss of fine-grained visual details; and the unbalanced representation of key visual concepts in the token space can result in limited recognition capability for certain visual patterns.

These methods differ significantly from tokenization strategies used in natural language processing. In the NLP domain, algorithms such as BPE form semantically meaningful tokens by combining frequently co-occurring character sequences \citep{galle2019investigating}, a strategy that has proven beneficial for Transformer learning \citep{makkuva2024attention, merrill2024language}. Research by \citep{rajaraman2024toward} demonstrates that for Markov sequence data, even smaller Transformer models can significantly reduce prediction loss when using appropriate tokenization strategies.

\subsection{BPE Tokenization for Visual Data}

Extending tokenization strategies such as BPE to the visual domain represents an emerging research direction \citep{razzhigaev2022pixel,chan2024analyzing}. The core idea is to form a more semantically meaningful visual vocabulary by combining frequently co-occurring visual patterns based on quantized images. Previous work \citep{zhang2025bpe} first proposed a theoretical framework for applying BPE to quantized visual data, demonstrating that for two-dimensional data conforming to specific generative processes, using visual BPE can significantly reduce model prediction loss. Their analysis indicates that compared to unigram models based on independent and identically distributed assumptions, BPE tokenization can more effectively capture structured patterns in visual data.

However, significant challenges remain in the transition from theory to practice. The implementation of \citep{zhang2025bpe} uses primarily frequency-based encoding without considering spatial relationships that are crucial in visual data. The relationship between vocabulary size and performance, as well as strategies that balance frequency with spatial consistency, remain underexplored. Besides, designing effective training procedures that leverage the unique properties of BPE visual tokens is also important. Research on parameter freezing strategies \citep{gu2024semantic, sun2024exploring} suggests that different settings at various training stages can impact the performance, but these findings have not been systematically applied to visual BPE training. Our work addresses these gaps by exploring enhancement strategies for visual BPE tokenization across vocabulary construction, data composition, and training procedures. Through these contributions, we aim to advance the practical application of BPE visual tokenization in multimodal foundation models.
\begin{figure*}[ht]
    \centering
    \includegraphics[width=.92\textwidth]{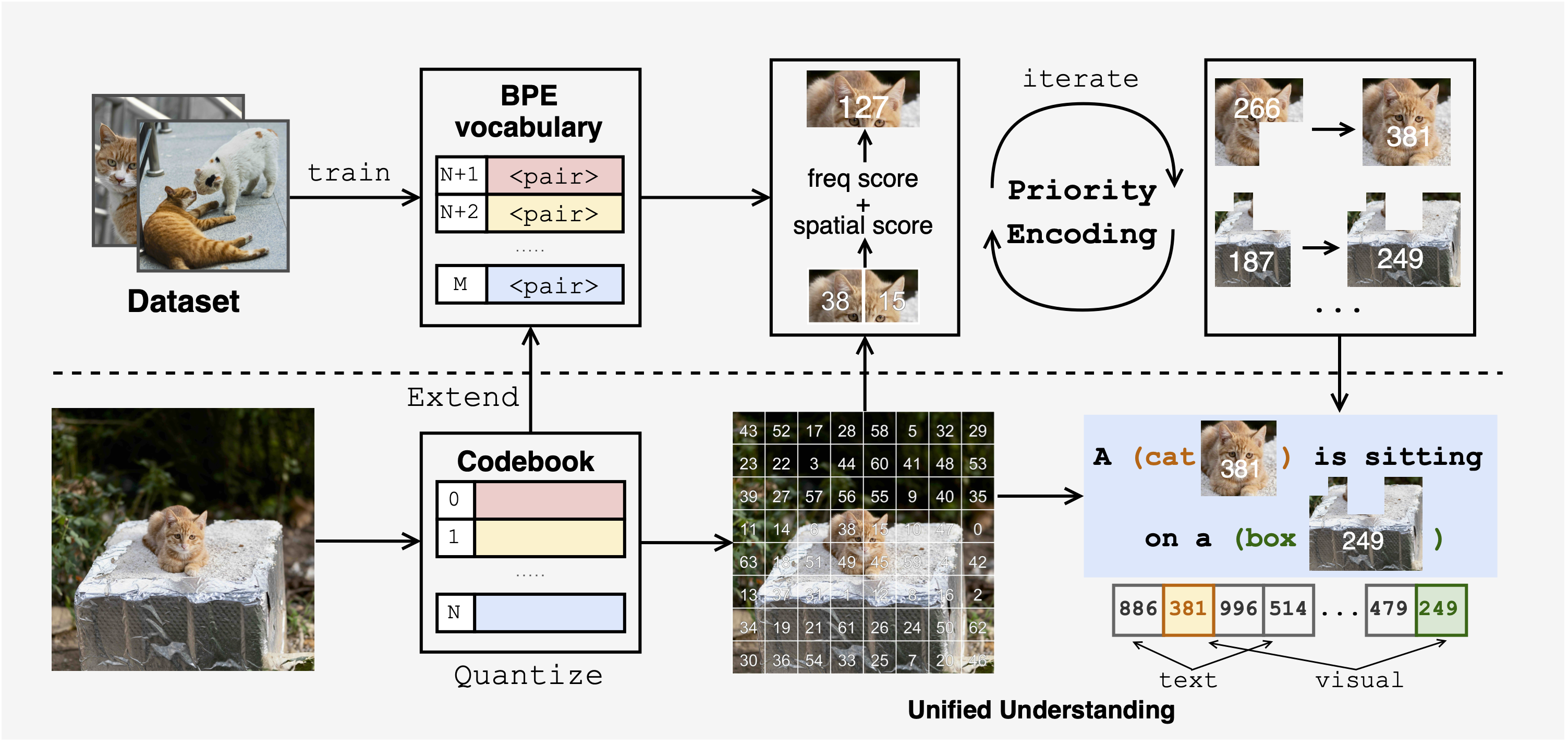}
    \caption{Overview of our framework. The upper part shows the BPE vocabulary construction process: starting from a training dataset, we identify candidate token pairs and apply our priority scoring mechanism (combining frequency and spatial consistency) to iteratively extend the vocabulary with new encoded tokens. The lower part illustrates the MLLM pipeline: an input image is first quantized using a VQ-GAN codebook, then encoded using the trained vocabulary. The resulting visual tokens (\eg, token \texttt{381} representing the whole ``cat'' and \texttt{249} representing the whole ``box'') are seamlessly integrated with text tokens to form a unified sequence for multimodal understanding.}
    \label{fig:framework}
\end{figure*}

\section{Method}
\label{sec:method}

In this section, we first begin by introducing key concepts and notation for our visual tokenization framework. Then we introduce the details of our method.

\subsection{Visual Tokenization Pipeline}

The core of our approach involves transforming visual information into token sequences that can be processed by language models. This transformation occurs through a pipeline with the following key steps:

\noindent \textbf{Vector Quantization (VQ).} 
Given an image $I$ divided into patches, we first apply a VQ model with a codebook $V = \{v_1, v_2, ..., v_K\}$ containing $K$ entries. This process maps each image patch to its closest codebook entry, producing a grid of discrete indices $Q(I) = \{q_{i,j}\}_{i,j=1}^{h,w}$, where $h$ and $w$ represent the height and width of the patch grid.

\noindent \textbf{BPE Visual Tokenization.} 
The BPE visual tokenization process, denoted as $T(Q(I))$, identifies frequently co-occurring patterns in the quantized representation and encodes them into new tokens, creating an extended vocabulary $D = V \cup V_{ext}$, where $V_{ext}$ represents the set of encoded tokens. Formally, the tokenization function $T: Q(I) \rightarrow (t_1, t_2, ..., t_n)$ maps the quantized image to a sequence of tokens $t_i \in D$, where $n \leq h \times w$ due to the encoding of multiple indices into single tokens. This approach parallels BPE tokenization in text processing but adapts for visual data, incorporating spatial relationships that are unique to the visual modality.

\subsection{Token-Based MLLM}

\noindent \textbf{Unified Token Sequence.} 
We formalize our token-based multimodal language model as a conditional probability distribution over token sequences. Given a text input $X = (x_1, x_2, ..., x_m)$ and an image input $I$, the model processes the combined sequence and generates output tokens autoregressively. The combined input consists of special tokens marking the beginning and end of the tokenized image representation:
\begin{equation}
S = (x_1, \ldots, x_m, [\mathrm{BOI}], t_1, \ldots, t_n, [\mathrm{EOI}],\ldots),
\end{equation}
where $[\mathrm{BOI}]$ and $[\mathrm{EOI}]$ are special tokens indicating the image boundary in the sequence.

\noindent \textbf{Autoregressive Modeling.} 
The model defines a probability distribution over the next token given all previous tokens:
\begin{equation}
p(s_i | s_{<i}) = \text{Softmax}(f_{\theta}(E(s_1), E(s_2), ..., E(s_{i-1}))),
\end{equation}
where $f_{\theta}$ represents the transformer-based language model with parameters $\theta$, and $E(s_i)$ represents the embedding of token $s_i$. For model generation, tokens are sampled sequentially:
\begin{equation}
\hat{s}_i \sim p(s_i | \hat{s}_1, \hat{s}_2, ..., \hat{s}_{i-1}, x_1, x_2, ..., x_m, t_1, t_2, ..., t_n).
\end{equation}

\noindent \textbf{Training Objective.} 
The model is trained to minimize the negative log-likelihood of the target sequence $Y = (y_1, y_2, ..., y_l)$ conditioned on the input:
\begin{equation}
\mathcal{L}(\theta) = -\mathbb{E}_{(X, I, Y) \sim \mathcal{D}} \left[ \sum_{i=1}^{|Y|} \log p_{\theta}(y_i | y_{<i}, X, T(Q(I))) \right],
\end{equation}
where $\mathcal{D}$ represents the training dataset containing input-output triplets.

\subsection{Framework Overview}

As formalized in the preliminaries above, our token-based multimodal model processes visual information through a pipeline that includes vector quantization $Q(\cdot)$ and visual tokenization $T(\cdot)$ before integration with text tokens in a language model. Our optimizations target each component of this pipeline while considering their interdependencies.

Specifically, our framework enhances this pipeline through: 
(1) Enhanced Vocabulary Construction; (2) Optimized Data Composition Strategies; (3) Efficient Multi-Stage Training. 
These components work together to improve the model's ability to process visual information, align visual and textual representations, and generate accurate responses to multimodal inputs. 

\begin{figure}[t]
    \centering
    \includegraphics[width=.59\linewidth]{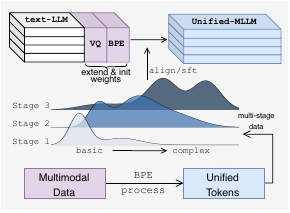}
    \caption{Illustration of our multi-stage training strategy. The bottom part shows multimodal data being processed into unified tokens. The middle curves represent the data distribution across training stages, showing a gradual shift from basic concepts (Stage 1) to more balanced (Stage 2) and finally complex reasoning-focused data (Stage 3). The top part illustrates how we expand a text-only LLM to incorporate visual capabilities through weight initialization and alignment, resulting in a unified multimodal model. This curriculum approach mirrors the hierarchical nature of BPE visual tokens, which progressively capture more complex visual patterns throughout the learning process.}
    \label{fig:data-composition}
\end{figure}

Figure \ref{fig:framework} illustrates the overall framework, which operates in two phases. In the vocabulary construction phase (upper part), we train a BPE vocabulary using our priority-guided encoding scheme that considers both co-occurrence frequency and spatial consistency. This approach creates an extended vocabulary that effectively captures meaningful visual patterns beyond what frequency-only methods can achieve. In the application phase (lower part), input images are first quantized through a VQ-GAN model, then encoded using the trained vocabulary. The resulting visual tokens are integrated with text tokens to form a unified sequence for the language model. Figure \ref{fig:data-composition} illustrates the model expanding and training process.

\subsection{Vocabulary Construction}

The effectiveness of token-based representations hinges on creating a vocabulary that efficiently captures meaningful patterns in the data. Our enhanced vocabulary construction method extends beyond simple frequency-based encoding to incorporate structural information specific to visual data, while relying exclusively on image statistics without reference to textual information.

In standard BPE tokenization, token pairs are encoded based solely on their co-occurrence frequency. While effective for text, this can be suboptimal for visual data where spatial relationships play crucial roles. We introduce a priority function $P(a, b)$ that guides the encoding process with a more comprehensive evaluation of candidate token pairs:
\begin{equation}
P(a, b) = F(a, b) + \alpha \cdot S(a, b).
\end{equation}

This function combines two key factors:

\begin{itemize}
    \item \textbf{Co-occurrence frequency} $F(a, b)$: The raw frequency with which tokens $a$ and $b$ appear adjacent to each other, normalized across the training corpus of quantized images.
    
    \item \textbf{Spatial consistency} $S(a, b)$: A measure of how consistently the token pair maintains spatial relationships across different images, calculated as:
    \begin{equation}
    S(a, b) = \frac{1}{N_{a,b}}\sum_{i=1}^{N_{a,b}} d(u_i(a, b), \bar{u}(a, b)).
    \end{equation}
\end{itemize}

\noindent Here, $u_i(a, b)$ represents the relative positioning of tokens $a$ and $b$ in instance $i$, $\bar{u}(a, b)$ is the average relative positioning across all instances, $N_{a,b}$ is the number of co-occurrences, and the spatial distance function is defined as:
\begin{equation}
d(u_1, u_2) = \exp\left(-\frac{\|u_1 - u_2\|^2}{2\sigma^2}\right),
\end{equation}
where $\|u_1 - u_2\|$ is the Euclidean distance between the two relative position vectors, and $\sigma$ is a scaling parameter that controls how quickly similarity decreases with distance. This formulation ensures that token pairs with consistent spatial relationships across images receive higher scores.

The weighting parameter $\alpha$ controls the relative importance of each factor and is determined through cross-validation on a development set. This dual-factor approach ensures that encoded tokens represent not only frequent patterns but also spatially consistent visual structures. It's important to emphasize that this entire process relies exclusively on image data, with no dependency on textual information or captions. The algorithm identifies patterns based solely on the statistical and structural properties of the visual content. Algorithm \ref{alg:priority_bpe} outlines our priority-guided encoding process with diversity filtering. For complete details, please refer to Algorithm \ref{alg:detailed_bpe} in the Appendix.

\begin{algorithm}[t]
\caption{Priority-Guided Encoding}\label{alg:priority_bpe}
\begin{algorithmic}
\State \textbf{Input:} Quantized training data $\mathcal{C}$, initial vocabulary $V$, target vocabulary size $|D|$
\State $D \gets V$
\While{$|D| < $ target size}
    \State Compute $F(a, b)$ and $S(a, b)$ for all adjacent token pairs in $\mathcal{C}$ 
    \State Calculate priority $P(a, b) = F(a, b) + \alpha \cdot S(a, b)$ and select top-$k$ pairs by priority: $\{(a_1, b_1), \ldots , (a_k, b_k)\}$.
    \State $(a^*, b^*) \gets \arg\max_{i \in \{1,...,k\}} P(a_i, b_i)$
    \State Create new token $c=(a^*, b^*)$
    \State $D \gets D \cup \{c\}$
    \State Update $\mathcal{C}$ by replacing all adjacent occurrences of $(a^*, b^*)$ with $c$ (horizontally or vertically).
\EndWhile
\State \Return $D$
\end{algorithmic}
\end{algorithm}

\subsection{Model Expanding}

Once we have constructed our enhanced visual vocabulary $D$, we need to expand the pre-trained language model to accommodate these new tokens. This expansion process serves as the bridge between our tokenization approach and the subsequent model training. Given a pre-trained language model $M$ with a text vocabulary $V_{\text{text}}$ and corresponding embedding weights $E_{\text{text}} \in \mathbb{R}^{|V_{\text{text}}| \times d}$ (where $d$ is the embedding dimension), we expand the embedding layer to incorporate the visual vocabulary $D$:
\begin{equation}
E_{\text{expanded}} = \begin{bmatrix} E_{\text{text}}, E_{\text{visual}} \end{bmatrix} \in \mathbb{R}^{(|V_{\text{text}}| + |D|) \times d},
\end{equation}
where $E_{\text{visual}} \in \mathbb{R}^{|D| \times d}$ represents the newly added embeddings for visual tokens. This expansion increases the model's vocabulary size from $|V_{\text{text}}|$ to $|V_{\text{text}}| + |D|$.

For the initialization of the newly added visual token embeddings, we adopt He initialization \citep{he2015delving}, \ie, $E_{\text{visual}}[i,j] \sim \mathcal{N}(0, \sqrt{2/d_{l}})$, where $d_{l}$ is the embedding dimension. This initialization helps maintain proper signal magnitude throughout the network, preventing vanishing or exploding gradients during the early stages of training.

The process is illustrated in the top part of Figure \ref{fig:data-composition}. We first extend a VQ token embedding corresponding to the VQ-GAN codebook, and then extend a BPE token embedding corresponding to the BPE vocabulary. The extended weights from these two components constitute the initialization of the visual part. By default, we standardize both dimensions of them to 8K, corresponding to an 8K VQ codebook and an 8K BPE vocabulary, respectively.

\subsection{Multi-Stage Training}

With the expanded model supporting both text and visual tokens, we now introduce our multi-stage training pipeline. Our approach combines strategic data composition with stage-specific parameter freezing to maximize the effectiveness of visual tokenization. Figure \ref{fig:data-composition} illustrates the overall process.

\subsubsection{Curriculum-Based Data Composition}
\label{sec:data-composition}

We develop a curriculum-based data composition strategy, which is designed to complement the BPE visual tokenization. Unlike traditional visual encoders that extract features in a single forward pass, BPE visual tokenization creates a hierarchical representation where encoded tokens incrementally capture more complex visual patterns. This hierarchical nature of BPE tokens requires a matching learning curriculum: tokens must first establish basic semantic meanings through simple visual-text associations before they can be effectively utilized for complex reasoning.

Specifically, we categorize our training data into four types based on complexity: Foundation Data (basic image-caption pairs), Perception Data (detailed visual attributes), Reasoning Data (complex visual QA), and Instruction Data (multi-turn interactions).
Details of all types of datasets can be found in the Appendix \ref{sec:ft-data-detail}.

We employ a curriculum that progressively shifts emphasis across training stages, formalized as a composition ratio $R_i(s) = w_i(s)/\sum_j w_j(s)$, where $R_i(s)$ represents the proportion of data type $i$ at stage $s$ with corresponding weight $w_i(s)$. Each stage employs distinct composition ratios, transitioning from foundation-heavy in early stages to instruction-focused in later stages. This curriculum ensures that visual tokens first establish basic meanings before tackling complex reasoning tasks.

This stage-specific curriculum ensures that the model first develops strong foundational visual understanding before focusing on more complex tasks, aligning with the parameter freezing strategy of each stage.

\subsubsection{Progressive Parameter Unfreezing}

Complementing our data curriculum, we implement stage-specific parameter freezing to control which components of the model are updated at each stage. The parameter update can be expressed as:
\begin{equation}
\theta_{s+1} = \theta_s - \eta \cdot M_s \odot \nabla_{\theta} \mathcal{L},
\end{equation}
where $\theta_s$ represents the model parameters at stage $s$, $\eta$ is the learning rate, $\nabla_{\theta} \mathcal{L}$ is the loss gradient, $\odot$ denotes element-wise multiplication, and $M_s$ is a binary mask that determines which parameters are updated during stage $s$.

Our training pipeline consists of three stages with different masks and objectives:

\textbf{Stage 1: Embedding Alignment} focuses exclusively on training the newly extended visual token embeddings while freezing the rest of the model. The mask is defined as $M_1[i] = 1$ if parameter $i$ belongs to new embeddings, $0$ otherwise. This alignment establishes basic visual-linguistic associations without disrupting pre-trained language capabilities.

\textbf{Stage 2: Selective Fine-tuning} unfreezes early transformer layers while keeping later layers frozen, with $M_2[i] = 1$ if parameter $i$ is in layers $1$ to $k$ ($25\%$ of total layers in our default implementation), $0$ otherwise. This enhances cross-modal interactions in the feature integration layers using a broader range of data types.

\textbf{Stage 3: Full Fine-tuning} unfreezes all parameters ($M_3[i] = 1, \forall i$) and emphasizes complex reasoning and instruction-following data. This stage refines the model's advanced capabilities while building upon the foundation established in earlier stages.










\section{Experiments}

\begin{table*}[ht]
\centering
\resizebox{\textwidth}{!}{
\begin{tabular}{l|c|c|cccccc}
\toprule[1.5pt]
\multirow{2}{*}{\textbf{Model}} & \multirow{2}{*}{\textbf{Params}} & \multirow{2}{*}{\textbf{Tokenizer}} & \multicolumn{6}{c}{\textbf{Benchmark}} \\
 & & & \textbf{VQAv2} & \textbf{MMBench} & \textbf{MME-P} & \textbf{SciQA-IMG} & \textbf{POPE} & \textbf{VizWiz} \\
\midrule[1pt]
\multicolumn{9}{c}{\textit{Continuous Embedding-Based Models}} \\
\midrule
InstructBLIP & 7B & Continuous & 75.2 & 38.3 & 1212.8 & 60.5 & 81.5 & 34.5 \\
LLaVA-1.5 & 7B & Continuous & 78.5 & 64.3 & 1510.7 & 66.8 & 85.9 & 50.0 \\
mPLUG-Owl2 & 8B & Continuous & 79.4 & 64.5 & 1450.2 & -- & 86.1 & 54.5 \\
HyperLLaVA & 7B & Continuous & 79.1 & 65.9 & 1481.2 & \textbf{70.4} & \underline{86.3} & 51.9 \\
ShareGPT4V & 7B & Continuous & 80.6 & \underline{68.8} & \textbf{1567.4} & 68.4 & -- & 57.2 \\
VILA-1.5 & 8B & Continuous & \underline{80.9} & \textbf{72.3} & -- & -- & 84.4 & \textbf{58.7} \\
LLaVA-Next & 7B & Continuous & \textbf{81.8} & 67.4 & \underline{1519.0} & \underline{70.1} & \textbf{86.5} & \underline{57.6} \\
\midrule[1pt]
\multicolumn{9}{c}{\textit{Discrete Token-Based Models}} \\
\midrule
Chameleon & 7B & Discrete & 56.2 & 37.3 & 1297.5 & -- & 78.2 & 46.4 \\
Being-VL-0 & 8B & Discrete & 60.6 & 44.0 & 1316.2 & 64.3 & 81.3 & 48.2 \\
Unified-IO-2 & 7B & Discrete & 79.4 & 71.5 & -- & \textbf{86.2} & \textbf{87.7} & -- \\
\rowcolor{gray!20} \textbf{\ours} \textit{w/o BPE} & 8B & Discrete & 54.3 & 38.2 & 1301.2 & 57.8 & 76.1 & 45.0 \\
\rowcolor{gray!20} \textbf{\ours} & 8B & Discrete & \underline{80.2} & \underline{71.8} & \underline{1525.8} & \underline{70.3} & 84.3 & \underline{57.4} \\
\rowcolor{gray!20} \textbf{\ours}$^+$ & 8B & Discrete & \textbf{80.6} & \textbf{72.1} & \textbf{1536.3} & 69.0 & \underline{86.0} & \textbf{57.8} \\
\bottomrule[1.5pt]
\end{tabular}
}
\caption{Performance comparison of different visual-language models across standard benchmarks. Models are grouped by tokenizer type (Continuous / Discrete). The best performance in each group is \textbf{bolded}, while second-best result in each group is \underline{underlined}.}
\label{tab:eval-main}
\end{table*}

In this section, we empirically evaluate our designed framework and training pipelines. Based on the methods described in Section \ref{sec:method}, we train our model \textbf{\ours} with an 8K extended vocabulary and a variant \textbf{\ours}$^+$ with a 16K extended vocabulary. We first describe our experimental setup, then present comprehensive results demonstrating the effectiveness of our approach compared to baselines. We further conduct detailed analyses of each component in our framework to validate our design choices. Considering the limited space, please refer to the Appendix \ref{sec:implementaition-detail} for complete details about our experiments.

\subsection{Experiments Setup}
\textbf{Evaluation Benchmarks:} We conduct evaluation on widely-used benchmarks \citep{li2024survey,li2024survey-1} that assess different aspects of visual understanding: VQAv2 \citep{goyal2017making} and VizWiz \citep{gurari2018vizwiz} for general visual question answering, MMBench \citep{liu2023mmbench} for comprehensive multimodal understanding, MME-P \citep{fu2024mme} for perception capabilities, SciQA-IMG \citep{lu2022learn} for scientific reasoning, and POPE \citep{Li-hallucination-2023} for hallucination evaluation.

\noindent\textbf{Baselines:} We primarily focus on open-source models with comparable parameters and similar training data scales to ensure that performance differences reflect architectural choices rather than scale advantages. We compare our \textbf{\ours} against two kinds of baselines: continuous embedding-based models and discrete token-based models. Continuous embedding-based models include InstructBLIP \citep{dai2023instructblip}, LLaVA-1.5/LLaVA-Next \citep{liu2023llava,liu2024improved}, mPLUG-Owl2 \citep{ye2023mplugowl, ye2023mplugowl2}, HyperLLaVA \citep{zhang2024hyperllava}, ShareGPT4V \citep{chen2023sharegpt4v}, and VILA-1.5 \citep{lin2023vila}. Discrete token-based models include Chameleon \citep{team2024chameleon}, Being-VL-0 \citep{zhang2025bpe}, and Unified-IO-2 \citep{lu2024unified}. We also compare `\textbf{\ours} \textit{w/o BPE}', which is a variant that completely removes the BPE part and retains only the base VQ token embedding. This is used to investigate the effectiveness of our framework's core BPE design.

\subsection{General Visual Understanding}

\begin{figure}[t]
    \centering
    \includegraphics[width=.7\linewidth]{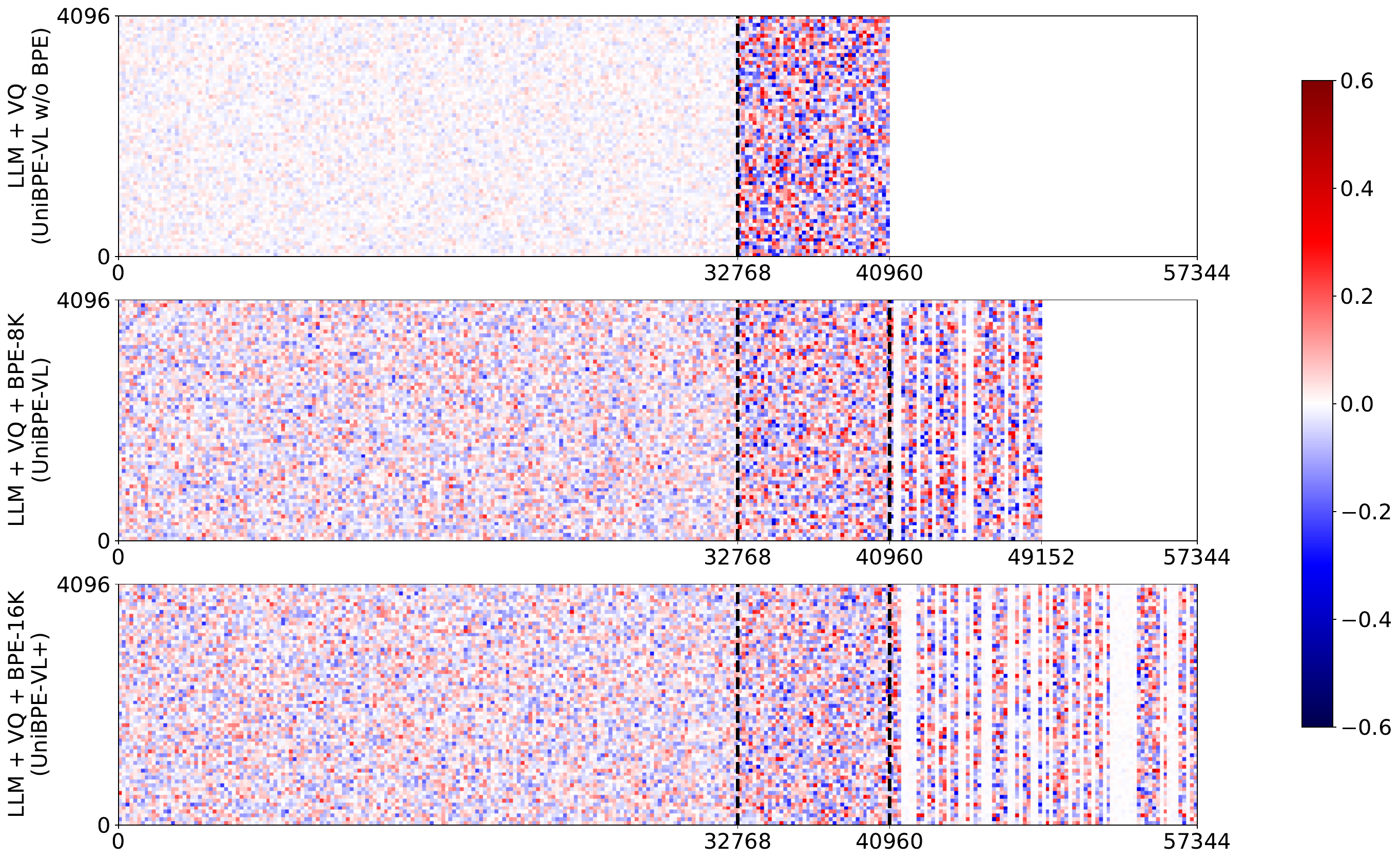}
    \caption{Visualization of embedding weight distributions across three model variants. The heatmaps represent sampled weights from the embedding matrices ($4096$ dimensions), wherein color intensity indicates weight magnitude (blue for negative, red for positive). The depth of the color indicates the activation magnitude of the corresponding token embedding during model inference. \textit{Top}: LLM+VQ (\textbf{\ours} \textit{w/o BPE}) with text tokens (32K) + VQ tokens (8K). \textit{Middle}: LLM+VQ+BPE-8K (\textbf{\ours}) with text tokens (32K) + VQ tokens (8K) + BPE tokens (8K). \textit{Bottom}: LLM+VQ+BPE-16K (\textbf{\ours}$^+$) with text tokens (32K) + VQ tokens (8K) + BPE tokens (16K).}
    \label{fig:embedding-weights}
\end{figure}

Table \ref{tab:eval-main} presents our models' performance compared to baselines. Our framework effectively narrows the performance gap between discrete token-based models and continuous embedding-based models. Both \textbf{\ours} and \textbf{\ours}$^+$ achieve competitive performance across all benchmarks, with \textbf{\ours}$^+$ reaching $80.6$ on VQAv2 and $72.1$ on MMBench, comparable to the best continuous embedding-based models like VILA-1.5 ($80.9$ and $72.3$ respectively). \textbf{\ours}$^+$ demonstrates strong perception capabilities with $1536.3$ on MME-P, outperforming many continuous embedding models.

Among discrete token-based approaches, Being-VL-0 represents the work most closely aligned with our method, as both implement BPE-based visual tokenization. Our method outperforms Being-VL-0 across all benchmarks, with notable gains on MMBench ($71.8$ \vs $44.0$) and MME-P ($1525.8$ \vs $1316.2$). These improvements are attributable to our enhanced vocabulary construction and optimized training strategies. Unified-IO-2 achieves stronger performance on SciQA-IMG ($86.2$ \vs our $70.3$), but it is worth noting that our models evaluate this benchmark in a zero-shot manner without task-specific fine-tuning. 

Additionally, our method achieves an optimal balance between capturing low-level visual details and modeling high-level semantic concepts, with reduced hallucination on POPE ($84.3$ for \textbf{\ours}, $86.0$ for \textbf{\ours}$^+$) compared to earlier discrete token approaches. When evaluating the overall performance across all benchmarks, our method demonstrates competitive results while maintaining the advantages of unified token representation.

Furthermore, the significant performance drop of `\textbf{\ours} \textit{w/o BPE}' on all benchmarks also highlights the necessity of the BPE vocabulary as a core part of our framework. We also conducted a qualitative case study, which provides additional evidence of \textbf{\ours}'s comprehensive visual understanding. For detailed results, please refer to Section \ref{sec:case-study}.

\subsection{BPE Token Activation Mechanism}

Figure \ref{fig:embedding-weights} visualizes embedding weight distributions across our model variants, revealing key insights into how BPE tokenization helps unifying multimodal knowledge. In the base LLM+VQ model (top), which is a variant removing BPE vocabulary, we observe a contrast between text (0-32K) and visual tokens (32K-40K). Text tokens exhibit predominantly low-magnitude weights (lighter colors), while visual tokens exhibit significantly higher magnitudes (deeper colors). This imbalance suggests that the model struggles to establish a unified representation space across modalities. With BPE tokenization (middle and bottom panels), this pattern changes substantially. These two models show more balanced weight distributions, suggesting BPE helps bridge the representation gap between modalities, creating a more unified semantic space for visual-textual interaction. This visualization elucidates why models with improved BPE patterns yield superior performance, thereby outperforming Being-VL-0.

However, some BPE tokens exhibit minimal activation (near-zero weights), appearing as white vertical stripes. This pattern is more evident in the 16K variant, where more embeddings remain unused. Intuitively, we hypothesize that this is because the current training does not fully utilize these 16K tokens, which may explain why \textbf{\ours}$^+$ underperforms \textbf{\ours} on certain tasks. Concurrently, this observation inspires us to examine the trade-offs between representational capacity and learning efficiency.

\begin{figure}[t]
    \centering
    \includegraphics[width=.7\linewidth]{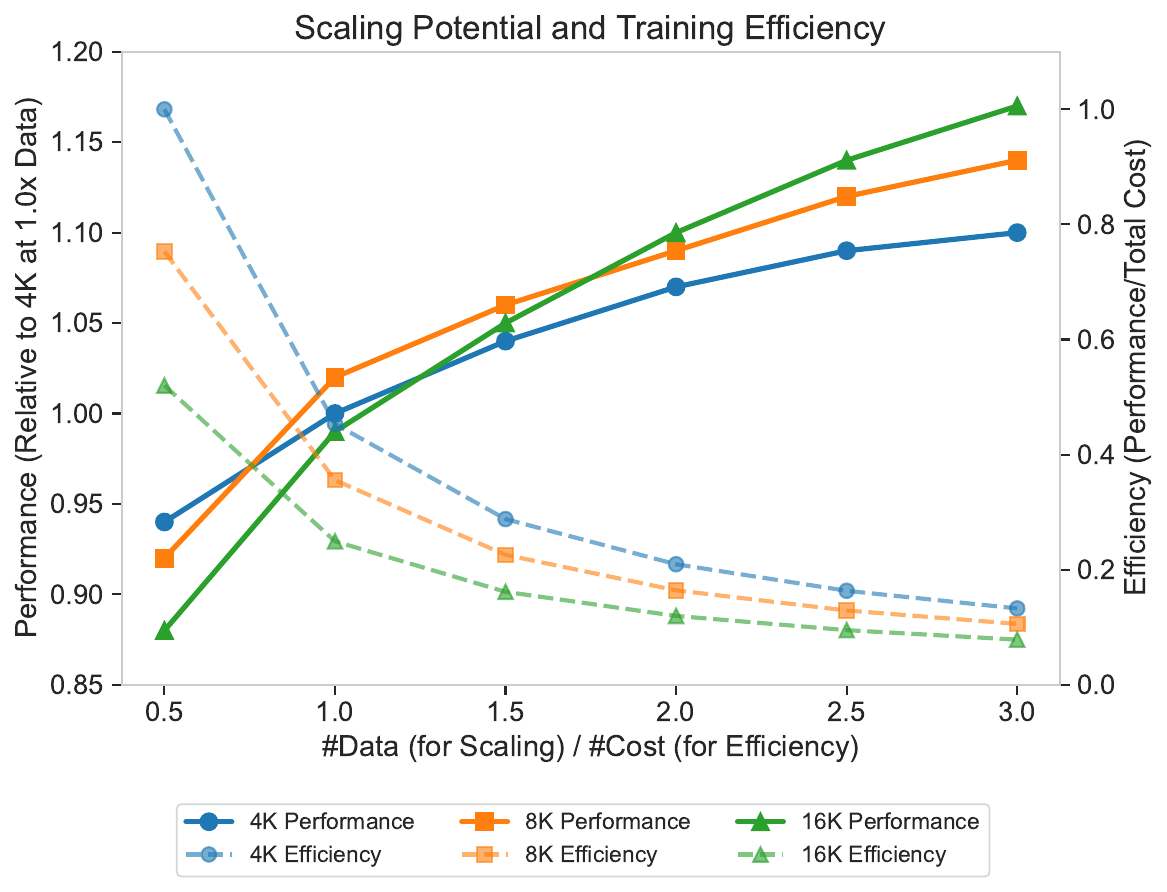}
    \caption{Scaling potential and training efficiency comparison for different vocabulary sizes (4K, 8K, 16K). Performance curves (solid lines) show relative performance gains as training data increases, while efficiency curves (dashed lines) illustrate the performance-to-cost ratio. All metrics are normalized relative to the 4K vocabulary model at standard training data amount ($1.0\times$).}
    \label{fig:scaling-efficiency}
\end{figure}

\subsection{Scaling Potential and Training Efficiency}

To further investigate the trade-offs between vocabulary size, performance scaling, and training efficiency, we conducted extended experiments using the same checkpoint. Figure \ref{fig:scaling-efficiency} presents our findings comparing three vocabulary sizes (4K/8K/16K) across varying data scales. For performance evaluation, we computed the normalized mean scores across four benchmarks (MME, POPE, VQAv2, and MMBench) and normalized them relative to the 4K vocabulary model at standard training data ($1.0\times$), providing an intuitive comparison of relative performance gains.

The horizontal axis serves dual purposes: for performance curves, it represents the relative amount of training data; for efficiency curves, it represents the total cost factor, incorporating computational resources $\times$ training time. Specifically, total cost increases with both vocabulary size and data amount. Our results reveal an important trade-off: while the 16K vocabulary shows superior scaling potential, eventually outperforming both 4K and 8K models as training data increases beyond $1.5\times$, its efficiency (performance-to-cost ratio) remains consistently lower than smaller vocabulary models. In contrast, the 8K vocabulary achieves a better balance, showing strong performance scaling while maintaining reasonable efficiency.

\begin{table}[t]
\centering
\begin{tabular}{l|cc|cc}
\toprule
\multirow{2}{*}{\textbf{Configuration}} & \multirow{2}{*}{\textbf{Curriculum}} & \multirow{2}{*}{\textbf{Progressive}} & \multicolumn{2}{c}{\textbf{Avg. Score}} \\
\cmidrule(lr){4-5}
 &  &  & \textbf{Perception} & \textbf{Reasoning} \\
\midrule
\rowcolor{gray!20} Standard & \checkmark & \checkmark & \textbf{80.3} & \textbf{71.1} \\
Progressive only & -- & \checkmark & 74.9 & 65.1 \\
Curriculum only & \checkmark & -- & 76.8 & 67.5 \\
Single stage & -- & -- & 71.2 & 62.3 \\
\bottomrule
\end{tabular}
\caption{Ablation study on different components in training. Configurations marked with ``--'' use a fixed data mixture ratio throughout training or employ a standard full fine-tuning approach.}
\label{tab:ablation-basic}
\end{table}

\subsection{Analysis of Multi-Stage Training}

We conduct ablation studies to investigate the impact of the multi-stage training. Table~\ref{tab:ablation-basic} shows that our standard approach achieves clear improvements over the single-stage training ($12.8\%$ and $14.1\%$ higher on perception and reasoning tasks respectively), demonstrating the effectiveness of the overall strategy. The results reveal that both curriculum-based data composition and progressive parameter unfreezing contribute to the performance, with curriculum data composition showing a stronger impact.

Table~\ref{tab:ablation-curriculum} shows that the shifting of data types matters significantly. Our curriculum design outperforms both reverse ordering and random mixing. This aligns with our design principle that visual BPE tokens should first establish basic semantic meanings before tackling complex reasoning tasks. Similarly, Table~\ref{tab:ablation-unfreezing} highlights the value of our three-stage progressive approach for parameter unfreezing. The simplified two-stage method shows modest performance differences ($2.7\%$ lower on perception tasks), while reverse unfreezing leads to more notable performance drops ($11.4\%$ lower on perception tasks), suggesting that the order of parameter updates affects how well the model integrates visual and textual information.

\begin{table}[ht]
\centering
\begin{tabular}{l|cc}
\toprule
\multirow{2}{*}{\textbf{Curriculum Strategy}} & \multicolumn{2}{c}{\textbf{Avg. Score}} \\
\cmidrule(lr){2-3}
 & \textbf{Perception} & \textbf{Reasoning} \\
\midrule
\rowcolor{gray!20} Progressive (FD $\to$ PD $\to$ RD $\to$ ID) & \textbf{80.3} & \textbf{71.1} \\
Reverse (ID $\to$ RD $\to$ PD $\to$ FD) & 73.4 & 64.0 \\
Random & 77.3 & 68.4 \\
\bottomrule
\end{tabular}
\caption{Analysis of different curriculum strategies. FD, PD, RD, and ID refer to Foundation Data, Perception Data, Reasoning Data, and Instruction Data as described in Section \ref{sec:data-composition}.}
\label{tab:ablation-curriculum}
\end{table}

\begin{table}[ht]
\centering
\begin{tabular}{l|cc}
\toprule
\multirow{2}{*}{\textbf{Unfreezing Strategy}} & \multicolumn{2}{c}{\textbf{Avg. Score}} \\
\cmidrule(lr){2-3}
 & \textbf{Perception} & \textbf{Reasoning} \\
\midrule
\rowcolor{gray!20} 3-stage (align $\to$ selective $\to$ full) & \textbf{80.3} & \textbf{71.1} \\
2-stage (align $\to$ full) & 78.2 & 69.3 \\
Reverse order & 72.1 & 63.0 \\
\bottomrule
\end{tabular}
\caption{Analysis of different parameter unfreezing strategies.}
\label{tab:ablation-unfreezing}
\end{table}
\section{Conclusions, Limitations, and Future Work}

This paper presents a unified visual tokenization framework that enhances multimodal understanding through three key innovations: priority-guided vocabulary construction that considers both frequency and spatial patterns, curriculum-based data composition that aligns with token complexity progression, and a specialized multi-stage training procedure. Our comprehensive experiments across diverse benchmarks demonstrate that this approach effectively bridges the gap between visual and textual modalities, achieving competitive performance compared to continuous embedding-based methods while maintaining the advantages of a unified token representation.

Our current study focused on models with 8B parameters due to computational constraints. While these models already show promising results, our scaling analysis indicates potential for further improvements with more training data. The observed relationship between vocabulary size and performance suggests that further scaling could achieve even stronger multimodal understanding capabilities.

A natural extension of our unified token-based approach is multimodal generation. Unlike models using separate encoders for different modalities, our framework enables the model to generate visual tokens in the same way it generates text tokens, establishing a truly unified representation paradigm. Although the current work concentrates on understanding tasks, extending our approach to incorporate generation capabilities constitutes a promising direction for future research.


\bibliographystyle{unsrt}
\bibliography{ref}

\clearpage

\beginappendix

\section{Implementation Details}
\label{sec:implementaition-detail}

\subsection{Model Configuration}

We use Llama-3.1-8B \citep{dubey2024llama} as our base language model due to its strong performance on language tasks and efficient architecture. For visual quantization, we employ a pretrained VQ-GAN model with a codebook size of 8192, which provides sufficient granularity for capturing visual details while maintaining computational efficiency. When applying our visual BPE tokenization, we experiment with extended vocabulary sizes ranging from 4K, 8K to 16K additional tokens to investigate the trade-off between representational capacity and learning efficiency as discussed in Section 4.3.

For priority-guided encoding, we set spatial consistency weight $\alpha=0.3$ and scaling parameter $\sigma=2.0$ by default, determined through ablation studies on a validation set. These parameters balance the influence of frequency and spatial consistency in token pair selection.

\subsection{Hyperparameters for the VQ-GAN model}

The hyperparameters for the VQ-GAN model used in our experiments are shown in Table \ref{tab:hyperparameters-vqgan}. The embedding dimension of 256 and codebook size of 8192 were chosen to provide sufficient representational capacity while maintaining computational efficiency. The input resolution of 512 allows for capturing fine-grained visual details without excessive memory requirements. We disabled dropout to preserve maximum visual information during the quantization process.

\begin{table}[ht]
\centering
\begin{tabular}{c|c}
\toprule[1pt]
\textbf{Hyperparameter} & \textbf{Value} \\
\midrule
embedding dimension & 256 \\
codebook size & 8192 \\
z\_channels & 256 \\
resolution & 512 \\
dropout & 0 \\
\bottomrule[1pt]
\end{tabular}
\caption{Hyperparameters for the VQ-GAN model}
\label{tab:hyperparameters-vqgan}
\end{table}

\subsection{Hyperparameters for multi-stage training}

The hyperparameters for multi-stage training are shown in Table \ref{tab:hyperparameters-train}. We carefully designed these parameters to align with the objectives of each training stage. In Stage 1, we use a higher learning rate (1e-3) to efficiently align the newly initialized visual token embeddings. For Stages 2 and 3, we reduce the learning rate (3e-5 and 5e-5 respectively) to prevent catastrophic forgetting while enabling meaningful updates to transformer layers. We increase gradient accumulation in later stages to effectively handle more complex data types. All stages use a cosine learning rate schedule with a 3\% warmup period to stabilize training.

\begin{table}[ht]
\centering
\begin{tabular}{c|c|c|c}
\toprule[1pt]
\textbf{Hyperparameter} & \textbf{Stage 1} & \textbf{Stage 2} & \textbf{Stage 3}\\
\midrule
batch size & 1 & 1 & 1\\
gradient accumulation & 2 & 4 & 4\\
learning rate & 1e-3 & 3e-5 & 5e-5\\
learning schedule & cosine & cosine & cosine \\
warmup ratio & 0.03 & 0.03 & 0.03 \\
weight decay & 0 & 0 & 0 \\
epoch & 2 & 3 & 3 \\
optimizer & AdamW & AdamW & AdamW\\
deepspeed stage & 2 & 2 & 2\\
\bottomrule[1pt]
\end{tabular}
\caption{Hyperparameters for multi-stage training}
\label{tab:hyperparameters-train}
\end{table}

\subsection{Data Details}
\label{sec:ft-data-detail}

Following our categorization in Section \ref{sec:data-composition}, we use a diverse set of datasets for different training stages.

\begin{itemize}
    \item \textbf{Foundation Data (FD)}: We use 595K images from CC-3M \citep{sharma2018conceptual}, 558K from LCS \citep{liu2024visual}, and a subset of LAION-2B-en \citep{schuhmann2022laion} for basic image-caption alignment.
    
    \item \textbf{Perception Data (PD)}: We incorporate 50.6K samples from RefCOCO \citep{kazemzadeh2014referitgame} and 66.2K from AOKVQA \citep{schwenk2022okvqa} to enhance detailed visual perception.
    
    \item \textbf{Reasoning Data (RD)}: We utilize 504K general QA entries and 343K reasoning-focused entries from the LLaVA-OneVision Dataset \citep{li2024llava}.
    
    \item \textbf{Instruction Data (ID)}: We include 57.3K entries from ShareGPT4o \citep{chen2023sharegpt4v}, 70K from ALLaVA Inst \citep{chen2024allava}, 180K OCR-related entries from LLaVA-OneVision, and 100K from Infinity-MM \citep{gu2024infinity}.
\end{itemize}

Our curriculum learning approach relies on carefully designed data composition ratios that shift across training stages:

\begin{itemize}
    \item \textbf{Stage 1 (Embedding Alignment)}: $R_{FD} \gg R_{PD} > R_{RD} = R_{ID} = 0$, focusing primarily on foundation data with some perception data
    
    \item \textbf{Stage 2 (Selective Fine-tuning)}: $R_{FD} \approx R_{PD} > R_{RD} > R_{ID}$, with increased emphasis on perception and reasoning data
    
    \item \textbf{Stage 3 (Full Fine-tuning)}: $R_{ID} > R_{RD} > R_{FD} \approx R_{PD}$, prioritizing instruction and reasoning data
\end{itemize}

Specifically, table \ref{tab:data-composition} presents the specific percentage breakdown for each data type across the three training stages.

\begin{table}[htbp]
\centering
\begin{tabular}{l|cccc}
\toprule
\textbf{Training Stage} & \textbf{FD} & \textbf{PD} & \textbf{RD} & \textbf{ID} \\
\midrule
Stage 1 (Embedding Alignment) & 80\% & 20\% & 0\% & 0\% \\
Stage 2 (Selective Fine-tuning) & 40\% & 30\% & 20\% & 10\% \\
Stage 3 (Full Fine-tuning) & 15\% & 15\% & 30\% & 40\% \\
\bottomrule
\end{tabular}
\caption{Data composition ratios (\%) across training stages}
\label{tab:data-composition}
\end{table}

\clearpage

\section{Case Study}
\label{sec:case-study}

In this section, we present qualitative examples to demonstrate the enhanced visual understanding capabilities of our approach compared to the frequency-only BPE baseline (Being-VL-0). These cases demonstrate several key advantages of our approach:

\begin{itemize}
    \item \textbf{Semantic Integrity}: Our priority-guided encoding better preserves complete semantic entities (people, animals, vehicles) as coherent token groups, enabling more accurate descriptions of subjects.
    
    \item \textbf{Spatial Relationship Understanding}: By incorporating spatial consistency in our encoding strategy, our model shows enhanced ability to describe relative positioning of elements within the scene.
    
    \item \textbf{Fine-grained Visual Detail Recognition}: Our approach better captures small but significant visual details, including colors, patterns, and distinctive features.
    
    \item \textbf{Structural Pattern Recognition}: The unified token created by our method facilitates stronger recognition of functional structures and their relationships within the scene.
\end{itemize}

\vspace{1cm}

\begin{center}
\textbf{\# Case 1}
\end{center}

\begin{center}
\fcolorbox{black}{white}{
\parbox{0.92\linewidth}{
\begin{minipage}[c]{0.2\linewidth}
\includegraphics[width=\linewidth]{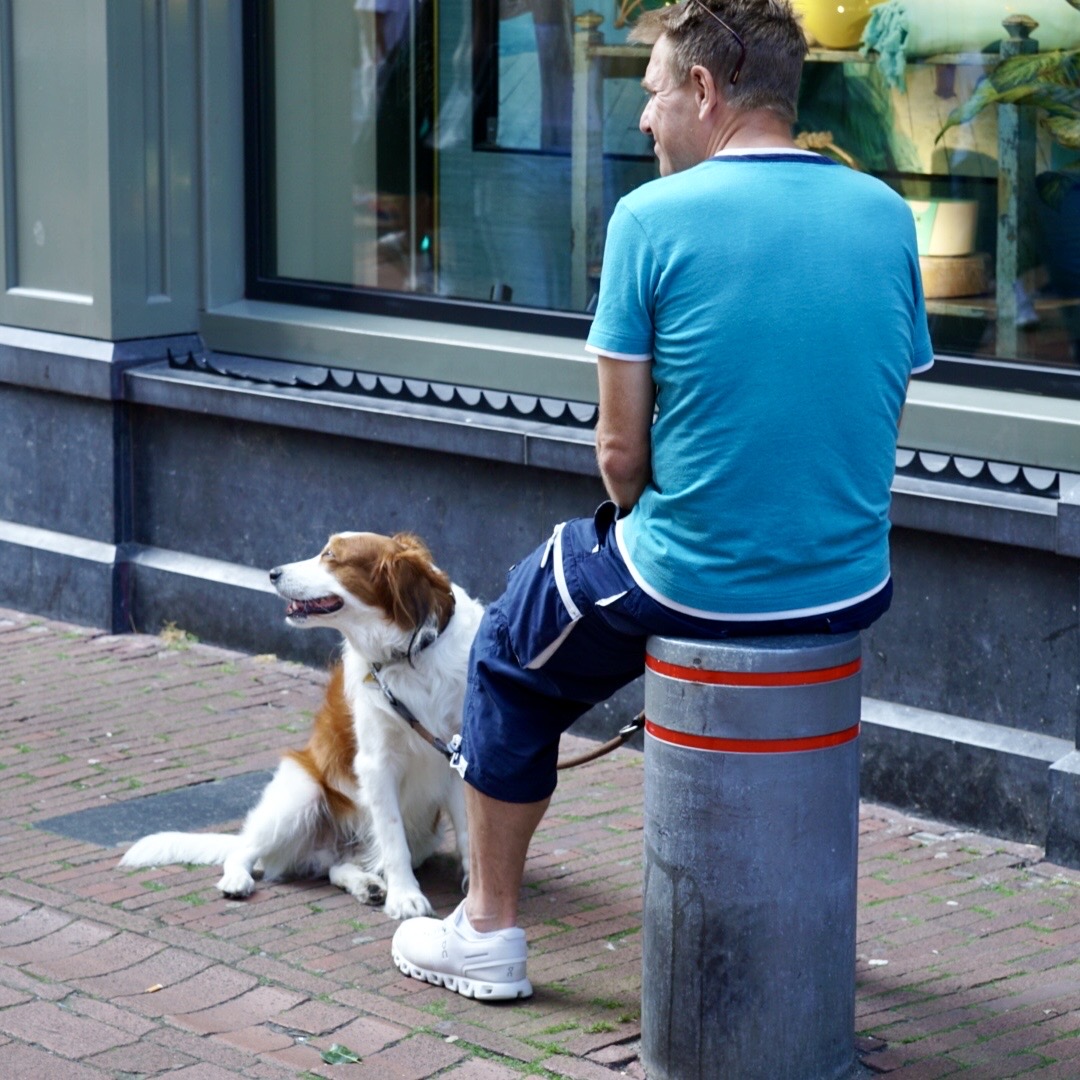}
\end{minipage}
\hfill
\begin{minipage}[c]{0.78\linewidth}
\small
\textbf{Question:} Please give me a detailed description of this image.
\end{minipage}
}}

\vspace{1mm}

\fcolorbox{black}{gray!10}{
\parbox{0.92\linewidth}{
\small
\textbf{\ours:} In this image, a man wearing a \textcolor{red}{turquoise t-shirt and navy shorts} is sitting on a \textcolor{red}{gray bollard with red stripes}. \textcolor{blue}{To his left, a brown and white dog sits on the brick pavement}, connected to the man by a leash. The dog appears calm while the man sits with one leg extended. They are on a brick sidewalk in front of a building with display windows.
}}

\vspace{1mm}

\fcolorbox{black}{white}{
\parbox{0.92\linewidth}{
\small
\textbf{Being-VL-0:} There is a person sitting with a dog beside him. The person is wearing a blue shirt and shorts. The dog is brown and white. They are in front of a building with windows. The person is sitting on a post, and the dog is sitting on the ground. This appears to be in a city, with a brick floor.
}}

\vspace{1mm}

\fcolorbox{black}{gray!10}{
\parbox{0.92\linewidth}{
\small
\textbf{Key Differences:} Our method captures \textcolor{red}{more precise visual details} (specific colors and features) and \textcolor{blue}{clearer spatial relationships} between the man and dog in the scene.
}}

\end{center}

\clearpage

\begin{center}
    \textbf{\# Case 2}
\end{center}

\begin{center}
\fcolorbox{black}{white}{
\parbox{0.92\linewidth}{
\begin{minipage}[c]{0.2\linewidth}
\includegraphics[width=\linewidth]{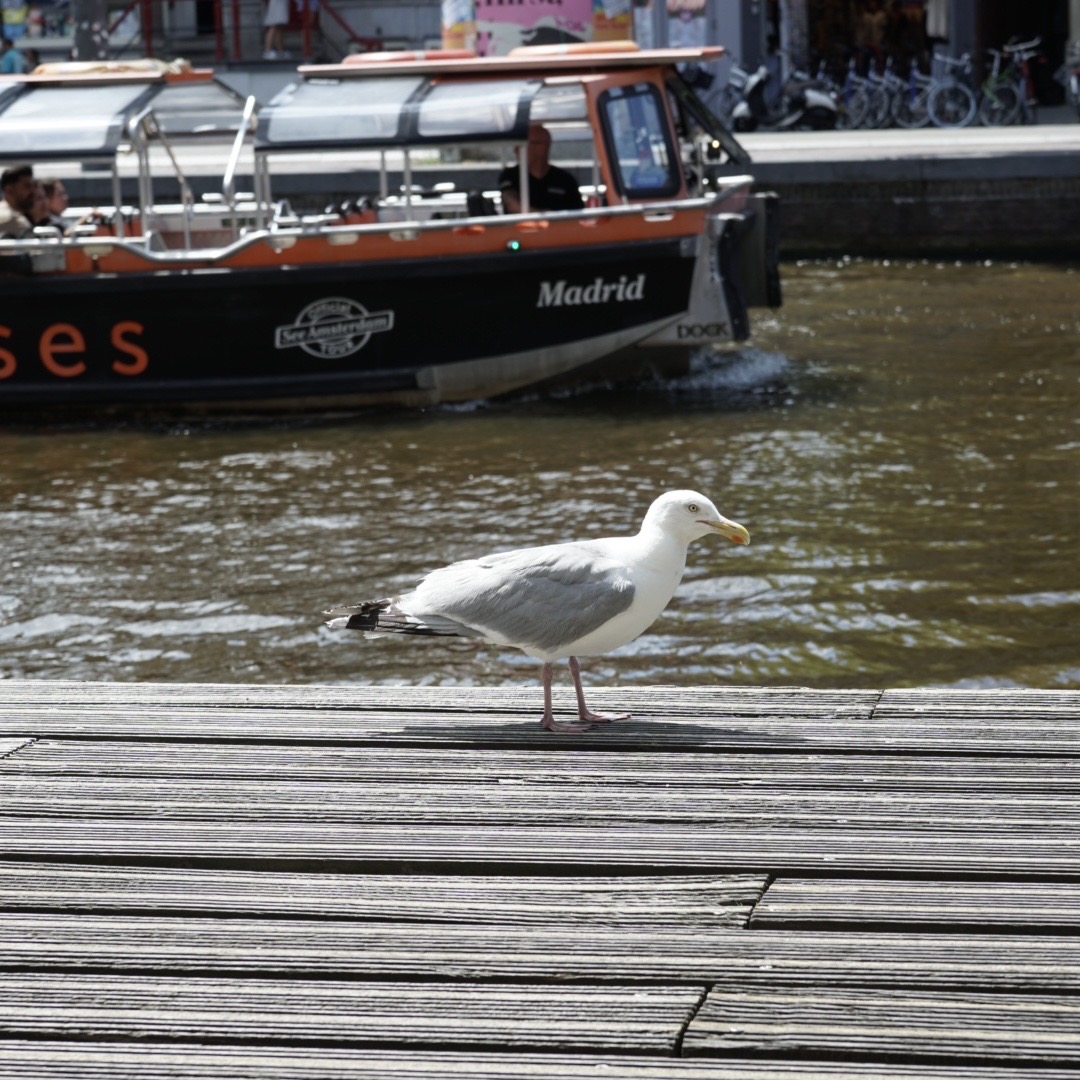}
\end{minipage}
\hfill
\begin{minipage}[c]{0.78\linewidth}
\small
\textbf{Question:} Please narrate what you observe in this image.
\end{minipage}
}}

\vspace{1mm}

\fcolorbox{black}{gray!10}{
\parbox{0.92\linewidth}{
\small
\textbf{\ours:} The image features a white seagull standing on wooden deck planks in the foreground. The bird is positioned in profile, facing right, with \textcolor{red}{its yellow beak and pink feet contrasting against its white and light gray feathers}. \textcolor{blue}{In the background is an orange and black tour boat navigating through brownish water}. The boat has a covered upper area and appears to be carrying passengers. The wooden deck seems to be part of a wharf or landing platform.
}}

\vspace{1mm}

\fcolorbox{black}{white}{
\parbox{0.92\linewidth}{
\small
\textbf{Being-VL-0:} In the image, there's a seagull standing on a wooden platform. The seagull is white with a yellow beak. In the background, there's a boat on brown water. The seagull is looking in one direction while standing on wooden boards. The boat is orange and black with some people on it.
}}

\vspace{1mm}

\fcolorbox{black}{gray!10}{
\parbox{0.92\linewidth}{
\small
\textbf{Key Differences:} Our method provides \textcolor{red}{more detailed descriptions of visual elements} (seagull's features) and \textcolor{blue}{better spatial understanding} of the scene components.
}}
\end{center}

\vspace{1cm}

\begin{center}
    \textbf{\# Case 3}
\end{center}

\begin{center}
\fcolorbox{black}{white}{
\parbox{0.92\linewidth}{
\begin{minipage}[c]{0.2\linewidth}
\includegraphics[width=\linewidth]{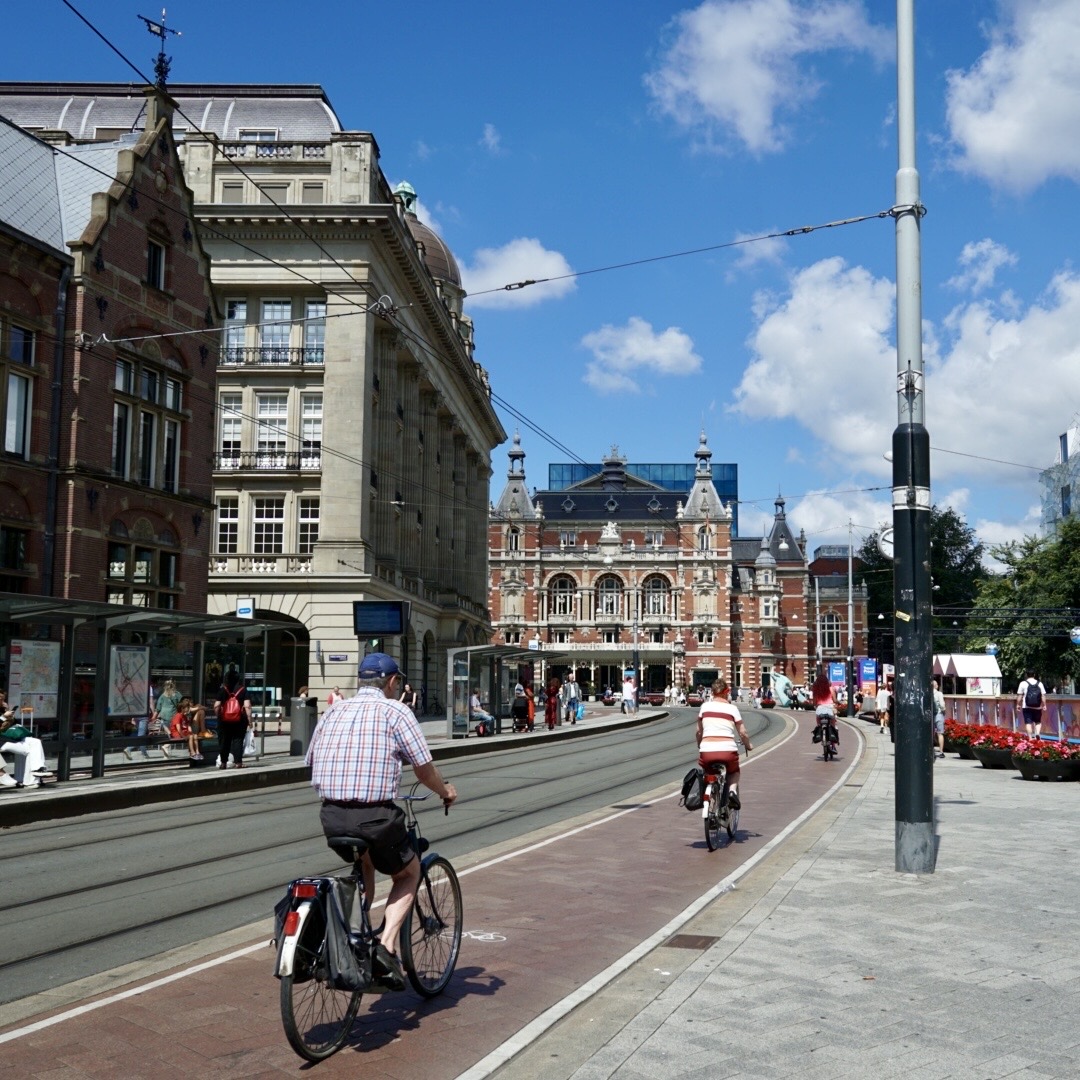}
\end{minipage}
\hfill
\begin{minipage}[c]{0.78\linewidth}
\small
\textbf{Question:} Could you describe what you see in this image?
\end{minipage}
}}

\vspace{1mm}

\fcolorbox{black}{gray!10}{
\parbox{0.92\linewidth}{
\small
\textbf{\ours:} This image shows a European urban transportation scene. In the foreground, \textcolor{red}{two cyclists are riding on a dedicated bike lane that runs parallel to tram tracks}. The leftmost cyclist wears a checkered shirt and has a bag on his bicycle rack. \textcolor{blue}{The scene displays structured transportation zones: pedestrian sidewalk, cycling path, and tram tracks integrated together}. In the background stands a \textcolor{red}{large historic building with distinctive architecture}. The sky is blue with white clouds.
}}

\vspace{1mm}

\fcolorbox{black}{white}{
\parbox{0.92\linewidth}{
\small
\textbf{Being-VL-0:} The image shows a city street with people riding bicycles. There are tram tracks on the ground and large buildings in the background. On the left, a person with a checkered shirt is riding a bicycle. The sky is blue with some clouds. This appears to be a European city based on the architecture and transportation setup.
}}

\vspace{1mm}

\fcolorbox{black}{gray!10}{
\parbox{0.92\linewidth}{
\small
\textbf{Key Differences:} Our method provides \textcolor{red}{more detailed description of key elements} and offers \textcolor{blue}{clearer understanding of the transportation infrastructure organization}.
}}
\end{center}

\clearpage

\begin{algorithm*}[ht]
\caption{Priority-Guided Encoding (Detailed Version)}\label{alg:detailed_bpe}
\begin{algorithmic}[1]
\State \textbf{Input:} Quantized training data $\mathcal{C}$, initial vocabulary $V$, target vocabulary size $|D|$, spatial weight $\alpha$, filtering threshold $\tau$
\State \textbf{Output:} Extended vocabulary $D$
\State $D \gets V$ \Comment{Initialize with base vocabulary}
\While{$|D| < $ target size}
    \State $P \gets \emptyset$ \Comment{Priority scores for token pairs}
    \For{each image $I$ in $\mathcal{C}$}
        \For{each position $(i,j)$ in $I$}
            \State Consider horizontal pair $(I_{i,j}, I_{i,j+1})$ if valid
            \State Consider vertical pair $(I_{i,j}, I_{i+1,j})$ if valid
            \State Update frequency counts for all considered pairs
        \EndFor
    \EndFor
    
    \For{each token pair $(a,b)$ with nonzero frequency}
        \State $F(a,b) \gets \text{count}(a,b) / \sum_{x,y} \text{count}(x,y)$ \Comment{Normalized frequency}
        
        \State $\bar{u}(a,b) \gets (0,0)$ \Comment{Initialize average relative position}
        \For{each occurrence of pair $(a,b)$ in position $(i,j,d)$}
            \State $u_i(a,b) \gets (0,1)$ if $d$ is horizontal, $(1,0)$ if $d$ is vertical
            \State $\bar{u}(a,b) \gets \bar{u}(a,b) + u_i(a,b)$
        \EndFor
        \State $\bar{u}(a,b) \gets \bar{u}(a,b) / N_{a,b}$ \Comment{Average relative position}
        
        \State $S(a,b) \gets 0$ \Comment{Initialize spatial consistency}
        \For{each occurrence of pair $(a,b)$ with position $u_i(a,b)$}
            \State $d(u_i, \bar{u}) \gets \exp(-\|u_i - \bar{u}\|^2 / 2\sigma^2)$ \Comment{Spatial similarity}
            \State $S(a,b) \gets S(a,b) + d(u_i, \bar{u})$
        \EndFor
        \State $S(a,b) \gets S(a,b) / N_{a,b}$ \Comment{Average spatial consistency}
        
        \State $P(a,b) \gets F(a,b) + \alpha \cdot S(a,b)$ \Comment{Combined priority score}
    \EndFor
    
    \State Select top-$k$ pairs by priority: $\{(a_1, b_1), \ldots, (a_k, b_k)\}$
    \State Filter out pairs with similarity $> \tau$ to existing tokens
    \State $(a^*, b^*) \gets \arg\max_{i \in \{1,\ldots,k\}} P(a_i, b_i)$
    \State Create new token $c=(a^*, b^*)$
    \State $D \gets D \cup \{c\}$
    \State Update $\mathcal{C}$ by replacing all adjacent occurrences of $(a^*, b^*)$ with $c$
\EndWhile
\State \Return $D$
\end{algorithmic}
\end{algorithm*}

\clearpage

\section{Broader Impact}

This work advances multimodal understanding through a unified token-based approach, with several potential societal implications. On the positive side, improved visual-language integration could enhance accessibility technologies for visually impaired users, enable more natural human-computer interaction, and support educational applications through better comprehension of multimodal learning materials. Our method's unified representation strategy may also lead to more computationally efficient models, potentially reducing the environmental footprint of multimodal AI systems.

However, like other powerful visual-language models, our approach could be misused to generate misleading content if deployed without proper safeguards. Models with enhanced visual understanding may also inherit or amplify biases present in training data. We encourage thoughtful consideration of these risks in downstream applications, including implementing appropriate content filtering, conducting fairness evaluations across diverse demographics, and establishing clear guidelines for responsible deployment. Furthermore, the growing computational requirements for training such models raise sustainability concerns that should be addressed through efficiency optimizations and responsible resource use.

\section{Detail of Priority-Guided Encoding}

Algorithm \ref{alg:detailed_bpe} presents the complete version of our priority-guided encoding process (Algorithm \ref{alg:priority_bpe}) in the main manuscript. The key extensions compared to the simplified version include: 

\begin{itemize}
    \item Comprehensive processing of both horizontal and vertical. adjacencies in two-dimensional visual data.
    \item Detailed calculation procedures for spatial consistency metrics
    \item Implementation of the diversity filtering mechanism to ensure vocabulary coverage.
\end{itemize}

\section{Licenses}

In our code, we have used the following libraries which are covered by the corresponding licenses:

\begin{itemize}
    \item Numpy (BSD-3-Clause license)
    \item PyTorch (BSD-3-Clause license)
    \item Transformers (Apache license)
    \item Numba (BSD-2-Clause license)
\end{itemize}

\clearpage

\end{document}